\documentclass[12pt,draftclsnofoot, onecolumn]{IEEEtran}
\usepackage{amsmath,amsfonts}
\usepackage[lined,boxed,commentsnumbered, ruled]{algorithm2e}
\usepackage{amssymb}
\usepackage{amsthm}
\usepackage{amsmath}
\usepackage{array}
\usepackage[caption=false,font=normalsize,labelfont=sf,textfont=sf]{subfig}
\usepackage{textcomp}
\usepackage{stfloats}
\usepackage{url}
\usepackage{verbatim}
\usepackage{graphicx}
\usepackage{cite}
\usepackage{bm}
\usepackage{xcolor}
 \def\BibTeX{{\rm B\kern-.05em{\sc i\kern-.025em b}\kern-.08em T\kern-.1667em\lower.7ex\hbox{E}\kern-.125emX}}

\definecolor{color1}{HTML}{D0B22B}
\definecolor{dred}{RGB}{128,0,0}
\definecolor{colorhkust}{RGB}{20,43,140}
\definecolor{colorshanghaitech}{RGB}{162,0,5}
\definecolor{colortsinghua}{RGB}{116,52,129}
\definecolor{colordark}{RGB}{184,134,11}
\usepackage{amsthm}
\theoremstyle{definition}

\newtheorem{theorem}{Theorem}

\newtheorem{definition}{Definition}
\newtheorem{remark}{Remark}
\newtheorem{assumption}{Assumption}

\newcommand{\transpose}{\mathsf{T}}
\newcommand{\Htranspose}{\mathsf{H}}
\newcommand{\norm}[1]{\left\|{#1}\right\|}

\newcommand{\expp}[1]{\mathbb{E}\left[{#1}\right]}

\SetKwProg{Forpara}{for each }{ do in parallel}{end}
\SetKw{Initialization}{Initialization:}

\begin{document}


\title{\LARGE \bf Federated Linear Bandit Learning via Over-the-air Computation}
\author{
	Jiali Wang, 
	Yuning~Jiang, 
	Xin Liu,
	Ting~Wang, 
	Yuanming Shi 
	\thanks{This work was supported in part by the National Key Research and Development Program of China (No. 2022ZD0119102). The work of Yuanming Shi was supported in part by the Natural Science Foundation of Shanghai under Grant No. 21ZR1442700, the National Nature Science Foundation of China under Grant 62271318, and the Shanghai Rising-Star Program under Grant No. 22QA1406100. The work of Yuning Jiang was supported in part by the Swiss National Science Foundation under the NCCR Automation (51NF40\_80545). }
	\thanks{Jiali Wang and Ting Wang are with MoE Engineering Research Center of Software/Hardware Co-design Technology and Application, and Shanghai Key Lab. of Trustworthy Computing, East China Normal University, China. {\tt 51215902015@stu.ecnu.edu.cn, twang@sei.ecnu.edu.cn}}	
	\thanks{Yuning Jiang is with Automatic Control Laboratory, EPFL, Switzerland. {\tt yuning.jiang@ieee.org}}
	\thanks{Xin Liu and Yuanming Shi are with School of Information Science and Technology (SIST), ShanghaiTech University, China. {\tt \{liuxin7, shiym\}@shanghaitech.edu.cn}}
	}

\maketitle

\begin{abstract}

In this paper, we investigate federated contextual linear bandit learning within a wireless system that comprises a server and multiple devices. Each device interacts with the environment, selects an action based on the received reward, and sends model updates to the server. The primary objective is to minimize cumulative regret across all devices within a finite time horizon. To reduce the communication overhead, devices communicate with the server via over-the-air computation (AirComp) over noisy fading channels, where the channel noise may distort the signals. In this context, we propose a customized federated linear bandits scheme, where each device transmits an analog signal, and the server receives a superposition of these signals distorted by channel noise. A rigorous mathematical analysis is conducted to determine the regret bound of the proposed scheme. Both theoretical analysis and numerical experiments demonstrate the competitive performance of our proposed scheme in terms of regret bounds in various settings.

\end{abstract}

\section{Introduction}

Multi-armed bandit (MAB) is a general framework for sequential decision-making and is widely used in various domains such as recommender systems \cite{li2010contextual} and advertisements \cite{li2010exploitation}. MAB learning provides a principal method to balance exploration and exploitation in the face of an uncertain environment to maximize the reward. 
Contextual bandit learning~\cite{li2010contextual} extends the classical MAB model by incorporating rewards that depend on both the context and the selected action. For example, in a recommender system, the user sequentially requests recommendations for which item to purchase next. The website can leverage additional information, such as past purchase records, browsing history, etc., to improve learning models. Contextual bandit learning takes advantage of the diversity in users' preferences. However, learning the preferences of each user individually can be challenging. 
To address this, contextual linear bandit learning has been proposed, which generalizes context across users. The reward is assumed to depend on the unknown linear function of the feature vector across all users, with each action mapped to a feature vector.

Empowered by various distributed edge devices and large-scale decentralized applications, federated learning allows cooperation between different entities to improve performance under the coordination of an edge server without sharing their local data. 
Recent works have investigated the decentralized MAB problem \cite{lalitha2021bayesian,yang2022distributed,bistritz2018distributed,DBLP:conf/aaai/ShiS21}. 
For instance, to investigate the cooperative stochastic multi-armed bandits problem, \cite{lalitha2021bayesian} adopted a message-passing algorithm.
\cite{yang2022distributed} developed two effective algorithms for the heterogeneous agents in the system, while \cite{bistritz2018distributed} considered a fully distributed MAB problem and proposed an algorithm for the resource allocation and collision scenarios. \cite{DBLP:conf/aaai/ShiS21} studied the problem that each agent has a heterogeneous local model. Some other works focus on the federated MAB problem under constraints of privacy \cite{zhou2023differentially,dubey2020differentially}.

Distributed optimization with noisy fading wireless channels is another active research area \cite{cao2022distributed,guo2020analog,xu2021learning,yang2022over}.
\cite{cao2022distributed} focused on distributed constrained online convex optimization over noisy channels via over-the-air computation (AirComp). \cite{guo2020analog} proposed an over-the-air aggregation technique for FL to improve the aggregation quality and accelerate the convergence speed. AirComp enables fast, low-latency data aggregation by leveraging the waveform superposition property of multiple access channels (MAC). To adapt the fading channel and reduce the channel distortion, \cite{xu2021learning} introduced a local learning rate optimization algorithm based on FedAvg. \cite{yang2022over} implemented second-order optimization over the noisy fading channel to achieve efficient FL, exploiting the AirComp technique.

However, all the aforementioned prior works have focused on federated bandit learning without channel distortion, and there is currently a lack of methods to handle federated bandit learning over noisy fading channels. 
To address this issue and achieve high performance with limited resources, we propose using AirComp \cite{nazer2007computation}, a technique that enables devices to transmit their model updates over wireless channels simultaneously~\cite{yang2020federated}.
By leveraging the superposition property of multiple access channels, AirComp enables fast aggregation and reduces the total communication rounds required for federated bandit learning. 
This approach has previously been used for online convex optimization in \cite{cao2022distributed}, which essentially inspires us to extend it to propose an efficient AirComp-based federated bandit learning framework over noisy fading channels.
Our key contributions are summarized as follows.

\begin{itemize}
\item  We propose a novel AirComp-based bandit learning framework that exploits the principles of federated bandits learning and leverages the waveform superposition property of multi-access channels to enable fast model aggregation. The utilization of AirComp significantly improves communication efficiency by enabling all devices to transmit the signal simultaneously.

\item We provide  rigorous theoretical bounds on the regret of the scheme, demonstrating that our proposed scheme achieves a high probability regret bound of $ \mathcal{O}(\sigma \sqrt{MTd}( \log (\gamma_{\max} / \gamma_{\min} + TL^2 / d / \gamma_{\min} ) )$, where $M$ and $T$ are the number of devices and total iteration rounds, respectively, and $d$ is the dimension of context. 

\item Extensive simulations are conducted to evaluate the effectiveness of the proposed scheme. Simulation results demonstrate that our scheme achieves a competitive performance in terms of regret, and we also investigate the impact of key system parameters on the overall regret.

\end{itemize}

\textit{Notations:} Throughout this paper, $\mathbb{C}$ and $\mathbb{R}$ denote the complex and real numbers sets. We denote the operations of transpose and conjugate transpose by the superscripts $ (\cdot)^\transpose $ and $ (\cdot)^\Htranspose $.  $ \expp{\cdot}$ is the expectation operator. $\langle\bm{x}, \bm{y} \rangle $ represents the inner product of vector $\bm{x}$ and $\bm{y}$. $ \textbf{I}_{d}$ is an identity matrix in $\mathbb R^{ d \times d} $. Any symmetric matrix $ \bm{X} $ is positive semi-definite (PSD) if $ \bm{y}\bm{X}^\transpose \bm{y} \leq 0$. The ellipsoid $ \bm{X} $-norm of vector $ \bm{y} $ is $ \norm{\bm{y}}_{\bm{X}} = \sqrt{\bm{y}^\transpose \bm{X} \bm{y}} $ for any PSD matrix $ \bm{X} $.

\section{System model}
\label{sys_model}
\subsection{Federated Linear Bandits}
\begin{figure}[htbp!]
\centering
\includegraphics[width=0.65\linewidth]{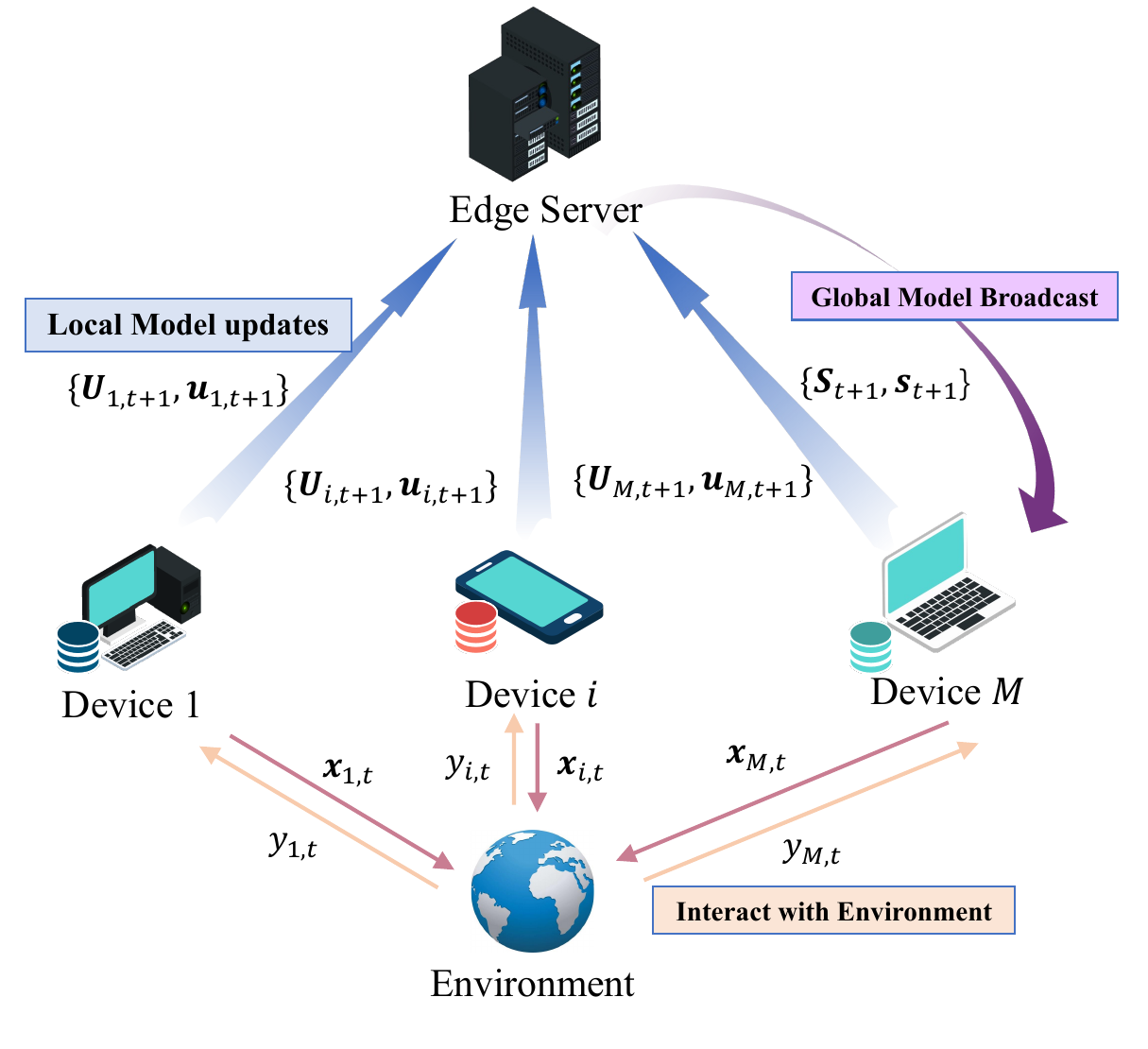}
\caption{Illustration of a wireless federated bandits learning system.}
\label{fig:sysmodel}
\end{figure}
This paper focuses on the federated linear bandits learning over wireless channels as shown in Fig.~\ref{fig:sysmodel}. We consider a typical wireless federated learning (FL) system with a single-antenna edge server and a set of $ M $ single-antenna devices. In the centralized architecture, devices communicate with the parameter server (PS) to train a model jointly over the wireless channel. At every iteration $ t\in[T] $, each device $ i\in[M] $ has its own decision set $ \mathcal{D}_{i,t} \subset \mathbb{R}^d $ and selects an action $ \bm{x}_{i,t} \in \mathbb{R}^d $. We assume that the reward $ y_{i,t} $ of device $ i $ at iteration $ t $ is linear with added sub-Gaussian noise, i.e., $  y_{i,t}=\bm{x}_{i,t} ^\transpose \bm{\theta}^* + \eta_{i,t} $, where $ \bm{\theta}^*\in \mathbb{R}^d$ and $ \eta_{i,t}  $ denote an unknown fixed parameter and a noise parameter, respectively \cite{dubey2020differentially}. 
Ideally, the goal of the devices is to minimize the cumulative group pseudo-regret \cite{dubey2020differentially}:
\begin{equation}
\mathcal{R}(T)=\sum_{i=1}^{M} \sum_{t=1}^{T} \langle\bm{x}_{i,t} ^*-\bm{x}_{i,t} , \bm{\theta}^*   \rangle,
\end{equation}
with optimal $\bm{x}_{i,t}^*=\arg\max_{\bm{x}\in\mathcal{D}_{i,t}}  \langle \bm{x}, \bm{\theta}^* \rangle$. Two sets of parameters are collected at each device, as shown in Alg.~\ref{alg1}. One is all observations up to the last synchronization round $ t^\prime $, and the other is its observations from the last synchronization round $ t^\prime $ to $ t $. Due to the influence of the wireless channel, these parameters are distorted by channel noise, i.e., the Gram matrix 
\begin{equation}
\bm{S}_{t^\prime+1} =\sum_{i=1}^M {\bm{U}}_{i,t^\prime+1} + \bm{N}_t,
\end{equation}
and the reward vector 
\begin{equation}
\bm{s}_{t^\prime+1}=\sum_{i=1}^M {\bm{u}}_{i,t^\prime+1} + \bm{n}_t ,
\end{equation}
where $  \bm{N}_t $ and $\bm{n}_t   $ denote the effective noise caused by the wireless channel. Thus, $ \bm{S}_t $ and $ \bm{s}_t $ can be rewritten as
\begin{equation}
\bm{S}_t =\sum_{i=1}^M\sum_{\tau=1}^{t^\prime} \bm{x}_{i,\tau}\bm{x}_{i,\tau}^\transpose + \bm{N}_{t^\prime},\bm{s}_t =\sum_{i=1}^M\sum_{\tau=1}^{t^\prime} y_{i,\tau}\bm{x}_{i,\tau} + \bm{n}_{t^\prime}.
\end{equation}
The final form of the parameters at device $ i $ at any iteration $t > t^\prime$, we have
\begin{equation}
\bm{V}_{i,t} = \sum_{\tau=t^\prime}^{t-1} \bm{x}_{i,\tau}\bm{x}_{i,\tau}^\transpose+\bm{S}_t, \;\tilde{\bm{u}}_{i,t}= \sum_{\tau=t^\prime}^{t-1} y_{i,t}\bm{x}_{i,\tau}+\bm{s}_t.
\end{equation}
The confidence ellipsoid is centered at the ridge regression estimator $ \bar{\bm{\theta}}_{i,t} = \bm{V}_{i,t}^{-1}  \tilde{\bm{u}}_{i,t},$ and the confidence set $\mathcal{E}_{i,t}$ for device $i$ in the iteration $t$ is $    \mathcal{E}_{i,t}=\{ \bm{\theta} \in\mathbb{R}^d | \| {\bm{\theta} - \bar{\bm{\theta}}_{i,t}} \|_{\bm{V}_{i,t}} \leq \beta_{i,t} \}.$
Then, each device solves the following problem at each iteration to select the action that maximizes the upper confidence bound (UCB):
\begin{equation}
\bm{x}_{i,t}=\arg\max_{\bm{x}\in\mathcal{D}_{i,t}}\left( \langle \bar{\bm{\theta}}_{i,t} , \bm{x}\rangle + \beta_{i,t}\norm{\bm{x}}_{\bm{V}_{i,t}^{-1}}\right).
\end{equation}

To reduce the total number of global updates over $ T $, we adopt the event-triggered communication in \cite{dubey2020differentially, Wang2020Distributed}. Specifically, the global update will be triggered if the following event is true for any device $ i\in[M]$:
\begin{equation}
\log \frac{\det(\bm{V}_{i,t} + \bm{x}_{i,t}\bm{x}_{i,t}^\transpose + (\gamma_{\max}-\gamma_{\min})\textbf{I}_d)}{\det(\bm{S}_{i,t})}
\geq \frac{D}{\Delta t_i},
\end{equation}
where $ \Delta t_i $ is the time since the last synchronization, $ D>0 $ is a constant as a threshold for the event trigger, and $\gamma_{\min}$ and $\gamma_{\max}$ are the constants depending on the channel noise. The global update is done over the wireless channel and thus is affected by the channel noise. After receiving the signals from the devices, the edge server aggregates all the signals to generate new model parameters.

The whole training process is presented in Alg.~\ref{alg1}. It consists of three stages: (i) \emph{local device computing}, where each device, at every iteration $ t $, interacts with the environment, computes a UCB on each action within the decision set $ \mathcal{D}_{i,t} $, and selects the action with the largest UCB. 
(ii) \emph{model aggregation}, where each device $ i $ checks whether the synchronization event is true. If true, all edge devices transmit their local updates to the server over the wireless channel, and the server aggregates and update the parameters. (iii) \emph{global model broadcast}, where the server disseminates the current model parameters to all devices through the wireless channel. Based on the proposed algorithm, efficient federated linear bandit learning can be achieved over the noisy wireless channel.
\begin{algorithm}[htbp!]
\caption{Federated Linear Bandit Learning via AirComp}
\label{alg1}
\LinesNumbered
\KwIn{$ \forall i, $ set $ \bm{S}_{i,1} \leftarrow \gamma_{\min}\textbf{I}_d  ,\bm{s}_{i,1} \leftarrow \bm{0},$ $ \bm{U}_{i,1} \leftarrow \bm{0}, {\bm{u}}_{i,1} \leftarrow \bm{0},\Delta t_i \leftarrow 0. $
}
\For{each iteration $ t\in[T] $}{
\For{each device $ i \in [M] $}{
$ \bm{V}_{i,t} \leftarrow\bm{S}_{i,t} +\bm{U}_{i,t}, \tilde{\bm{u}}_{i,t} \leftarrow \bm{s}_{i,t} + {\bm{u}}_{i,t}  $\;
Receive the decision set $ \mathcal{D}_{i,t}$\;
Compute regressor $ \bar{\bm{\theta}}_{i,t} \leftarrow \bm{V}_{i,t}^{-1}  \tilde{\bm{u}}_{i,t}  $\;
Construct a confidence set $\mathcal{E}_{i,t}$ and compute confidence-set bound $ \beta_{i,t} $\;
Select $ \bm{x}_{i,t} \leftarrow \underset{\bm{x} \in \mathcal{D}_{i,t}}{\text{argmax}}\;\langle \bm{x},  \bar{\bm{\theta}}_{i,t}  \rangle  + \beta_{i,t} \norm{\bm{x}}_{\bm{V}_{i,t}^{-1}  }$\;
Observe the reward $ y_{i,t} $ for action $ \bm{x}_{i,t} $\;
Update 
$\bm{U}_{i,t+1} \leftarrow \bm{U}_{i,t} + \bm{x}_{i,t}\bm{x}_{i,t}^\transpose, \bm{u}_{i,t+1} \leftarrow \bm{u}_{i,t} + \bm{x}_{i,t} y_{i,t}$\;}
\eIf{$ \log \det(\bm{V}_{i,t} + \bm{x}_{i,t}\bm{x}_{i,t}^\transpose + (\gamma_{\max}-\gamma_{\min})\mathbf{I}_d)-\log\det(\bm{S}_{i,t}) \geq D/\Delta t_i $}{
Devices send $ \bm{U}_{i,t+1} $ and $ \bm{u}_{i,t+1} $ to the server through wireless channel via AirComp\;
Server receives the estimated aggregated signal $\bm{S}_{t+1}$ and $  {\bm{s}}_{t+1}  $ with distortion over the wireless channel via \eqref{noise_vector}\;
Server broadcasts $ {\bm{S}}_{t+1} $ and $ {\bm{s}}_{t+1} $ to all devices over error-free channels\;
Devices update $ {\bm{S}}_{i,t+1} \leftarrow {\bm{S}}_{t+1}$ and ${\bm{s}}_{i,t+1} \leftarrow {\bm{s}}_{t+1}  $\;
$\Delta t_i \leftarrow \Delta t_i + 1$ \;
}{
$ \bm{S}_{i,t+1} \leftarrow  \bm{S}_{i,t}$, $\bm{s}_{i,t+1} \leftarrow  \bm{s}_{i,t}$,  $\bm{U}_{i,t+1} \leftarrow  \bm{0}$, $\bm{u}_{i,t+1} \leftarrow  \bm{0}$, $\Delta t_i \leftarrow 0$\;
}
}
\end{algorithm}

\subsection{Communication Model}
\label{sub:communication}
In this subsection, we mainly consider transmitting local information from the devices to the edge server over multi-access fading channels in the uplink. We assume the downlink transmission is reliable and error-free due to a greater transmit power at the edge server than at the devices \cite{amiri2020machine,amiri2020federated}.

As the transmission of information through wireless channels confronts great challenges such as channel noise, limited resources, and high latency, we apply a technique called over-the-air computation \cite{nazer2007computation} to decrease the communication cost. Specifically, AirComp, as a non-orthogonal multiple access scheme, exploits the waveform superposition property of multiple access channels to enable all edge devices to perform simultaneous transmission. Furthermore, the work in \cite{yang2020federated} states that AirComp realizes fast and spectrum-efficient model aggregation for wireless FL and thus decreases communication overheads. 

In this paper, AirComp is adopted by all edge devices to transmit information to the edge server with perfect synchronization. We assume a block flat-fading channel, where each block is divided into $ K$ time slots for a $ K $-dimensional local model, i.e., $K=(d^2+3d)/2$. The channel coefficients during a block remain
unchanged. In this way, the model aggregation process can be performed over the air by transmitting $\bm{U}_{i,t}$ and $\bm{u}_{i,t}$ in a single coherent block. Without loss of generality, we only describe one case that transmits $\bm{u}_{i,t}  $ in detail here, and the other case that transmits $  \bm{U}_{i,t}$ is quite similar, i.e., we pick entries on or above the main diagonal of $  \bm{U}_{i,t} $ to transmit based on the fact that $  \bm{U}_{i,t} $ is symmetric. More specifically, given $  \alpha_{i,t} $ as the precoder of device $ i $ in the iteration $ t $, the received signal at the server can be represented as follows:
\begin{equation}
\hat{\bm{s}}_{t}=\sum_{i=1}^{M} h_{i,t}	\alpha_{i,t} \bm{u}_{i,t} + \tilde{\bm{n}}_t,
\end{equation}
where $ h_{i,t}\in\mathbb{C} $ is the channel coefficient for  device $ i $ in the $ t $-th iteration,  $ \beta_t $ is the denoising factor at the edge server, and $ \tilde{\bm{n}} _t\in\mathbb{R}^d \sim \mathcal{CN}(0, \sigma_n^2\textbf{I}_d) $ is the additive white Gaussian noise (AWGN) vector. The transmit power constraint in iteration $ t $ at device $i$ can be given by
\begin{equation}
\expp{\norm{ \alpha_{i,t} \bm{u}_{i,t} }_2^2} \leq d \times P_0,
\end{equation}
where $ P_0 $ denotes the transmit power and the signal-to-noise (SNR) can be expressed as ${P_0}/{\sigma_n^2}$.
In addition, we assume that perfect channel state information (CSI) is available at all edge devices and the server as in most of the existing works on AirComp \cite{yang2020federated,sery2020analog,amiri2020machine}. To alleviate the influence of the additive noise and channel fading and improve the performance of AirComp, designing power control policies for magnitude alignment and signal processing schemes is essential. Each device implement channel inversion and set $ \alpha_{i,t} $ as
\begin{equation} \label{denou}
\alpha_{i,t} =\sqrt{\rho_t} \frac{(h_{i,t})^\Htranspose}{|h_{i,t}|^2},
\end{equation}
where $\sqrt{\rho_t}  $ denotes the denoising factor at the edge server. With designed $ \alpha_{i,t} $, the estimated signal at the edge server can be given by
\begin{equation} \label{noise_vector}
\hat{\bm{u}}_t=\frac{\hat{\bm{s}}_{t}}{\sqrt{\rho_t}} = \sum_{i=1}^{M} \bm{u}_{i,t} + \bm{n}_t,
\end{equation}
where $ \bm{n}_t \sim \mathcal{N}(0, \frac{\sigma_n^2}{\rho_t}\textbf{I}_d )$ is the effective noise vector. For simplicity, we assume that $  \sigma_t^2  \triangleq\frac{\sigma_n^2}{\rho_t}$. An imperfect estimation over the wireless channel causes an aggregation error during the training procedure. It is natural to jointly tune the transmit scalar and the denoising factor $ \sqrt{\rho_t} $ in each iteration to reduce the error gap. Based on the technique of channel inversion, which is commonly used in many existing works \cite{zhu2019broadband,zhu2020one}, the denoising factor $ \rho_t $ can be set as
\begin{equation}
\rho_t = \min _{i\in[M]} \frac{\|h_{i,t}\|^2dP_0}{\norm{\bm{u}_{i,t}}_2^2}.
\end{equation}
With this setting, the signal recovered by the server is an unbiased estimation of the transmitted information \cite{yang2020federated,nazer2007computation }.
          
\section{Theoretical Regret Analysis }\label{analysis_sec}
This section first recaps some preliminary assumptions and definitions for federated linear bandit learning. Then, we provide a regret analysis for Alg.~\ref{alg1} using the communication model presented in~Section~\ref{sub:communication}. 

\subsection{Preliminaries}
This subsection introduces some key assumptions and definitions for federated linear bandit learning. 
\begin{assumption}\label{ass1}
The action set is bounded by $ \norm{\bm{x}_{i,t}} \leq L $ and the decision set $ \mathcal{D}_{i,t}$ is compact $,\forall i\in[M],t\in[T] $.
\end{assumption}
\begin{assumption}\label{ass2}
The mean reward is bounded by $ \forall \bm{x}, \langle\bm{\theta}^*, \bm{x}\rangle  \leq 1$ and the target parameter is bounded by $ \norm{\bm{\theta}^*} \leq S $.
\end{assumption}
\begin{assumption}\label{ass3}
The rewards are bounded by $|y_{i,t}|\leq B$ and the noise parameter in the reward $ \eta_{i,t} $ is $\sigma$-sub-Gaussian for all $i\in[M]$, $t\in[T]$. \end{assumption}
Assumption~\ref{ass1}-\ref{ass3} are widely adopted in federated linear bandits learning \cite{Wang2020Distributed,dubey2020differentially,pmlr-v31-agrawal13a}. To establish meaningful regret bound and have control over the noise, we construct the spectral bounds on $ \bm{N}_t $ and $ \bm{n}_t $ as follows:
\begin{definition}\label{def_1} (Bounds for the channel noise)
Consider a subsequence $ [T]=1,\dots,T $ of size $ n $, a random symmetric matrix $( \bm{N}_{t} )_{ t\in[T]}$ with each (upper-triangle) entry drawn i.i.d. from Gaussian distribution $ \mathcal{N}(0, \sigma_t^2) $, and a random vector $(\bm{n}_{t})_{t\in[T]} $ whose entries are also drawn i.i.d. from $ \mathcal{N}(0, \sigma_t^2) $.
If for each round $ t\in[T] $ and device $ i $, the bounds are $ (\alpha/2nM) $-accurate for $( \bm{N}_{t} )_{t\in[T]}$ and $(\bm{n}_{t})_{t\in[T]} $:
\begin{align}\label{norm_N}
\norm{ \bm{N}_{t} } \leq& C\sigma_t \sqrt{d} \log(\frac{1}{\alpha}) \triangleq \gamma_{\max} ,\\\label{norm_N-1}
\norm{ \bm{N}_{t}^{-1} } \leq & 1/  \underbrace{\left( \frac{\alpha-2\exp(-Cd)}{2c}  \sigma_t   (\sqrt{d} - \sqrt{d-1})\right)}_{\gamma_{\min}},\\\label{norm_n_N}
\norm{ \bm{n}_{t} }_{  \bm{N}_{t} ^{-1}} \leq& \sqrt{2C\norm{\bm{N}_t^{-1}} \norm{\bm{n}_t}^2} =\kappa
\end{align}
with probability at least $ 1-\alpha/2nM $, where $ 0\leq \rho_{\min} \leq \rho_{\max} $, $\kappa>0$, and $C,c$ are constants. $\gamma_n$ is defined as $ \norm{\bm{n}_t} \leq \sigma_t \left(\sqrt{d} + \sqrt{2\ln \left(\frac{2}{\alpha} \right)}\right) \triangleq \gamma_{n}$\cite{sheffet2015private}.
\end{definition}
\begin{remark}
Inequality \eqref{norm_N} results from \cite{sheffet2015private}. Inequality \eqref{norm_N-1} and \eqref{norm_n_N} come from \cite{rudelson2009smallest} and \cite{zajkowski2020bounds}, respectively. Note that we assume that $ \bm{N}_t $ is a symmetric and non-negative definite matrix for simplicity since we can add $ r\cdot\textbf{I}_{d\times d} $ to $ \bm{N}_t $ to transform the output into a positive definite matrix according to \cite{zajkowski2020bounds}, where $ r=\expp{\norm{\bm{N}_t}} $. For the practical wireless system, the server first checks the received signal and then performs postprocessing to obtain the desired signals.
\end{remark}

To construct the UCB, an exploration sequence for each device can be defined as follows:
\begin{definition}\label{def_2}
A sequence $ \beta_{i,t} , i\in[M], t\in[T]$ is $ (\alpha, M,T) $-accurate for $ \bm{N}_t $ and $ \bm{n}_t $, if it satisfies $ \norm{\bar{\bm{\theta}}_{i,t} - \bm{\theta}^*} _{\bm{V}_{i,t}} \leq \beta_{i,t}$ with probability at least $1-\alpha $ for the overall iterations simultaneously. Based on \cite{dubey2020differentially}, all $\beta_{i,t}$ can be upper-bounded by
\begin{equation}\label{bar_beta}
    \bar{\beta}_t =   \sigma \sqrt{2\log \frac{2}{\alpha} +d \log \left(  \frac{\gamma_{\max}}{\gamma_{\min}} + \frac{tL^2}{d\gamma_{\min}}\right) }+ S\sqrt{\gamma_{\max}} + \kappa.
\end{equation}
\end{definition}
According to Definition~\ref{def_2}, we obtain confidence-set bounds on the suitable exploration sequence $ \beta_{i,t} $ for each agent to form the UCB clearly.

\subsection{Regret Analysis}
In this subsection, we introduce bounds on the maximum regret of the proposed algorithm.

\begin{theorem}\label{theo1}
Suppose synchronization occurs in at least $ n=\Omega(d \log (\gamma_{\max} / \gamma_{\min} + TL^2/d \gamma_{\min})) $ rounds, with probability at least $ 1-\alpha $ the pseudo-regret of Algorithm 1 is bounded by
\begin{equation}
\mathcal{R}(T) \leq4\nu \bar{\beta_T}  \sqrt{ 2MTd \log_\nu\left( \frac{\gamma_{\max}}{\gamma_{\min}} + \frac{TL^2 }{d\gamma_{\min}}\right) + 1},
\end{equation}
where $\bar{\beta_t}$ is defined in \eqref{bar_beta}.
\end{theorem}

The proof is based on two key Definitions aforementioned and motivated by \cite{dubey2020differentially} and \cite{DBLP:conf/nips/ShariffS18}. More details refer to Appendix~\ref{theo1_proof}.

\begin{remark}
According to Theorem~\ref{theo1}, we choose the threshold $ D=2T d(\log (\gamma_{\max} / \gamma_{\min} + TL^2/d \gamma_{\min})+1) ^{-1} $ to achieve the regret bound $\mathcal{O}(\sigma \sqrt{MTd}( \log (\gamma_{\max} / \gamma_{\min} + TL^2 / d / \gamma_{\min} ) )$ (we set $\nu = e$) \cite{dubey2020differentially}. Theorem~\ref{theo1} investigates the relationship between the number of iterations and the regret bound for a fixed variance of channel noise via the dependence on the bounds of noise. Our observations are summarized as follows:
\begin{itemize}
    \item For time horizon $T$, $\mathcal{R}(T)$ is $\mathcal{O}(\sqrt{T}\log T)$, which implies as $T$ increases, the total regret also becomes larger. A large dimension $d$ and the number of participated devices $M$ may also incur large regret. The result matches the regret bound of stochastic linear bandits without Aircomp noise in \cite{lattimore2020bandit}.
    
    \item For the variance of channel noise, characterizing by $\gamma_{\min}, \gamma_{\max}$ and $\kappa$, Theorem~\ref{theo1} also implies that the channel noise has a big impact on the regret bound. Specifically, a very large effective channel noise variance leads to larger accumulative regret since the received information at the server is greatly perturbed by the channel condition.
\end{itemize}

\end{remark}


\section{Numerical Experiments}\label{exp_sec}
\begin{figure*}[htbp!]
\centering
\begin{minipage}{0.4\linewidth}
\centering
\includegraphics[width=\linewidth]{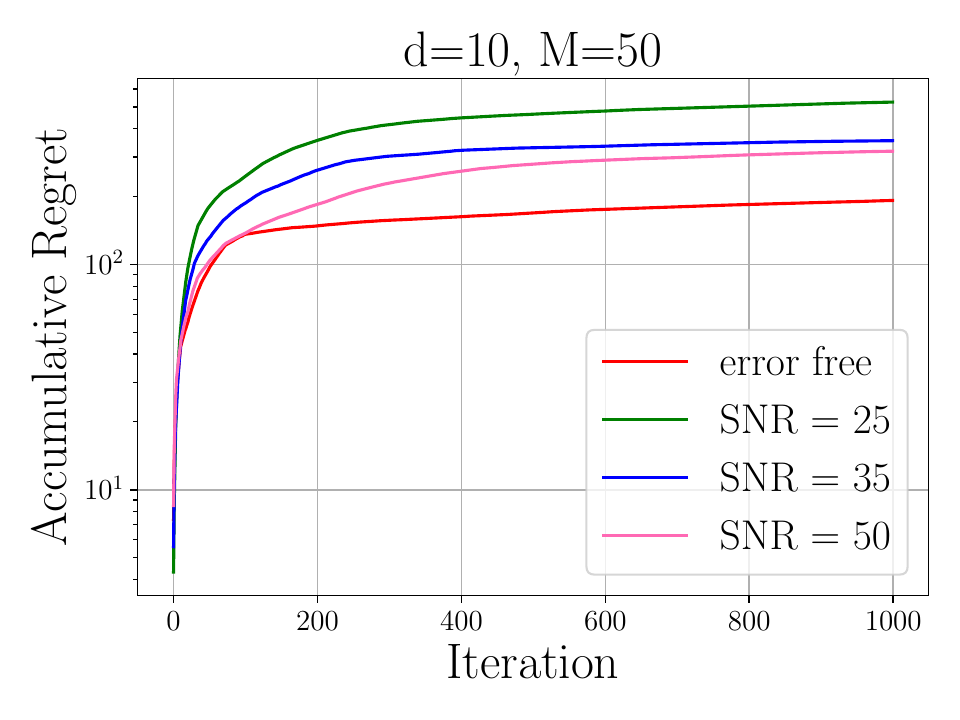}
\caption{The effect of different SNRs in regret.}
\label{fig:diff_noise}
\end{minipage}
\qquad
\begin{minipage}{0.4\linewidth}
\centering
\includegraphics[width=\linewidth]{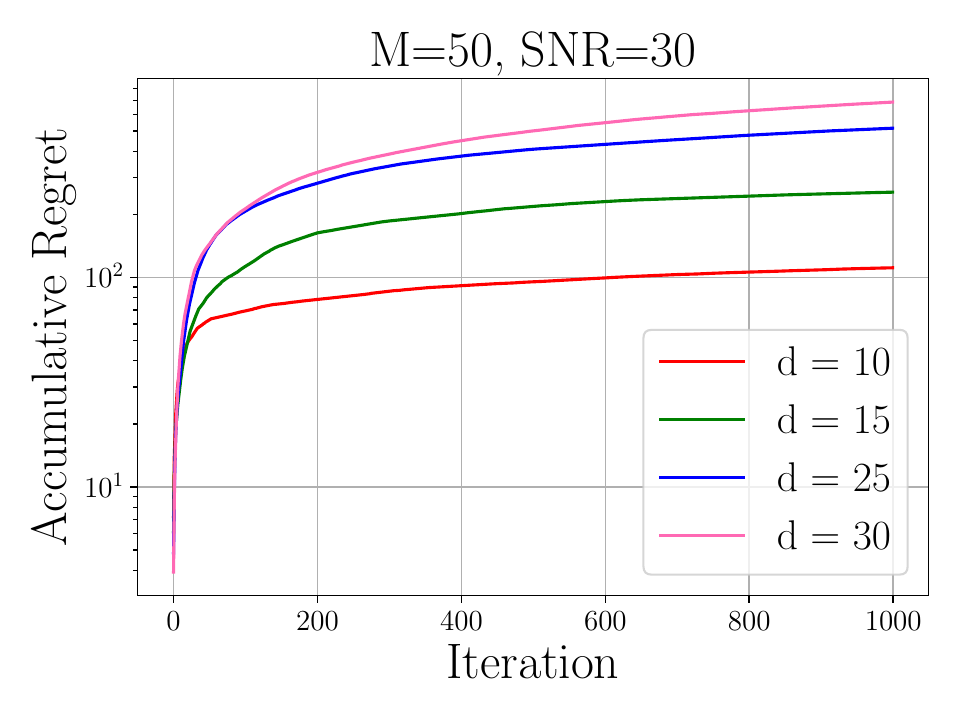}
\caption{The effect of different dimension $ d $.}
\label{fig:diff_d}
\end{minipage}
\begin{minipage}{0.4\linewidth}
\centering
\includegraphics[width=\linewidth]{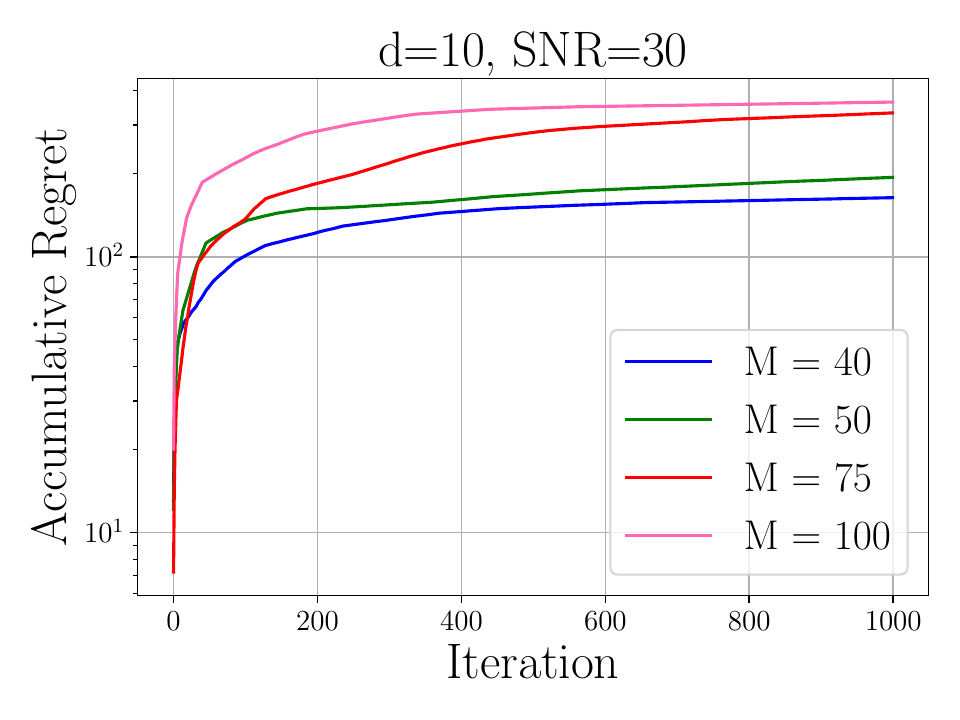}
\caption{The effect of different number devices.}
\label{fig:diff_m}
\end{minipage}
\end{figure*} 

In this section, we conduct extensive simulations to evaluate the performance of Alg.~\ref{alg1} over fading noisy channels. In the following, we first introduce the simulation settings, verify the performance of the proposed scheme and investigate the impacts of various key system parameters.
\subsection{Simulation Settings}
\subsubsection{Communication Settings}
We consider that a federated bandits learning system consists of a single-antenna server and $  M=50$ single-antenna devices. Each device is uniformly distributed in a circle area of radius $ R=500 $ m. The channel gain from each device to the edge server follows Rayleigh fading and the path loss model is given by
\begin{equation}
L(d)=\sqrt{G_0}\left(\frac{k}{k_0}\right)^{-\zeta},
\end{equation}
where $ k $ denotes the distance between the device $ i $ and the server, $ G_0 =10^{-3.35}$ is the average channel power gain with the distance to the server corresponding to $ k_0=1 $ m, and $ \zeta=2 $ is the path loss exponent factor. The channel coefficients are the small-scale fading coefficients $ h_{i,t}^{\prime} $ multiplied by the pass loss gain $ L(d) $, i.e., $ h_{i,t}=L(d) h_{i,t}^{\prime} $, $ h_{i,t}^{\prime} \sim \mathcal{C}\mathcal{N}(0,1)$. In our experiments, the maximum transmit power for each device is set to be $ P_0=23 $ dBm, the SNR is set to $ 80 $ dB. 
\subsubsection{Federated Bandit Learning Settings}For the federated bandits learning, we assume $ L=S=1 $ and fix $ \bm{\theta}^*\in \mathcal{B}_d(1) $. We randomly sample $ K\leq d^2 $ actions $ \bm{x} $ where for $ K-1 $ actions we have $ 0.5\leq\langle\bm{x}, \bm{\theta}^* \rangle \leq 0.6 $ and for the optimal we assume $ 0.7\leq\langle\bm{x}^*, \bm{\theta}^* \rangle \leq 0.8  $. The reward $ y_{i,t} $ is sampled from $ \text{Bernoulli}(\langle \bm{x}_{i,t}, \bm{\theta}^* \rangle) $ where satisfies $ \expp{y_{i,t} }=\langle \bm{x}_{i,t}, \bm{\theta}^* \rangle$ and $ |\langle \bm{x}_{i,t}, \bm{\theta}^* \rangle| \leq 1 $ \cite{dubey2020differentially}. We set $ T=1000 $ and $ d=10 $, and all experiments are averaged on $ 100 $ trials.

%
%

\subsection{Impacts of Key Parameters}
In this subsection, we verify the regret performance of the proposed scheme in terms of different key parameters, i.e. SNR, the dimension $d$, and the number of devices in the wireless system. 

In Fig.~\ref{fig:diff_noise}, we plot accumulative regret versus iteration for different SNR with fixed $ d=10, N=50 $. We consider Rayleigh fading
channels with three values for SNR, including $ 25$ dB, $35$ dB, and $50 $ dB. The error-free case presents perfect aggregation without channel noise. The error-free case reaches the best performance since there is no channel distortion during the training process. The other three cases of wireless fading channels demonstrate larger regrets. It is observed that the case with lower SNR achieves worse regret performance due to the stronger impact of channel noise through the training process. This is because the lower SNR is, the more errors would be introduced in the training procedure.

Fig.~\ref{fig:diff_d} illustrates accumulative regret versus the number of total iteration $ T $ by varying the dimension $ d $ with fixed the number of devices $M=50  $ and $ \text{SNR} = 30 $ dB. We focus on the effect of $ d $ in the overall regret over the noisy fading channels. A quadratic dependence in terms of dimension $ d $ can be observed in Fig.~\ref{fig:diff_d}. The result represents that for larger dimension $ d $, the total regret is also larger, which confirms the regret bound in Theorem~\ref{theo1}. 

In Fig.~\ref{fig:diff_m}, we numerically evaluate the total regret of the federated linear bandits problem under different number of devices $ M $ with fixed dimension $ d=10 $ and $ \text{SNR}=30 $ dB. It is observed that the accumulative regret increases as the number of devices increases. We observe a quadratic dependence over $ M $. More involved devices in the federated bandit learning may incur larger regret. The result is again consistent with the previous regret analysis in Theorem~\ref{theo1}.

\section{conclusion} \label{con_sec}


To address the challenge of signal distortion over noisy fading channels and facilitate efficient wireless data aggregation, this paper presented a novel AirComp-based federated linear bandit learning scheme.
Specifically, by leveraging the signal superposition property of multiple access channels, our AirComp-based approach significantly improved communication efficiency with low overhead, while our customized federated linear bandits method enabled unbiased transmission of model updates over noisy fading multiple access channels. Moreover, we provided a rigorous theoretical analysis of the accumulative regret bound of our proposed approach. Extensive numerical experimental results confirmed the effectiveness and efficiency of our approach.

\bibliographystyle{IEEEtran}
\bibliography{reference} 

\appendix
\section{proof of theorem 1}
\label{theo1_proof}
Let the sequence $ \bm{x}_{1,1},  \bm{x}_{2,1}, \dots,\bm{x}_{1,2}, \bm{x}_{2,2}, \dots,\bm{x}_{M,T}, $ denote the $ MT $ actions taken by a hypothetical device and $ \bm{W}_{i,t}  = \gamma_{\min}\textbf{I} + \sum_{j=1}^{M} \sum_{\tau=1}^{t-1}  \bm{x}_{j,\tau}\bm{x}_{j,\tau}^\transpose   + \sum_{j=1}^{i-1} \bm{x}_{j,t}\bm{x}_{j,t}^\transpose $ denote the Gram matrix obtained until the hypothetical device receive $ \bm{x}_{i,t} $. According to \cite{dubey2020differentially}, For any $ T_{k-1} \leq t\leq T_k $, the immediate pseudo-regret for any device $ i $ can be given by
\begin{equation}
\begin{aligned}
r_{i,t} &\leq 2 \bar{\beta_T} \norm{\bm{x}_{i,t}}_{\bm{V}_{i,t}^{-1}}\\
&\leq 2 \bar{\beta_T} \norm{\bm{x}_{i,t}}_{\bm{W}_{i,t}^{-1}} \sqrt{ \frac{\det(\bm{W}_{i,t})}{\det(\bm{G}_{i,t} + \gamma_{\min}\textbf{I})}}\\
&\leq 2\nu \bar{\beta_T} \norm{\bm{x}_{i,t}}_{\bm{W}_{i,t}^{-1}} .
\end{aligned}
\end{equation}
We assume that $ T_1, T_2, \dots, T_{\rho-1} $ denotes the trials when synchronization occurs. Consider performing $ T_k $ rounds of synchronization, the cumulative Gram matrices of all observations after $ T_k $ rounds can be defined as $ \bm{V}_k=\sum_{i=1}^{M} \sum_{t=1}^{T_k} \bm{x}_{i,t}\bm{x}_{i,t}^\transpose + \gamma_{\min}\textbf{I} $ and we have $ \bm{V}_0= \gamma_{\min}\textbf{I} $. The total regret can be divided into two parts, i.e., $R(T)=R(T,E) +R(T,\bar{E})$ where $E$ is the event to be the period $k$ when $1\leq \det(\bm{V}_k) / \det(\bm{V}_{k-1}) \leq \nu$ holds. Then we present how to bound these two items individually.
Summing up the immediate pseudo-regret over all such periods where $ E $ holds, with probability at least $ 1-\alpha $ the total regret can be expressed as
\begin{equation}\label{ine1}
\begin{aligned}
R(T,E) &= \sum_{i=1}^{M} \sum_{t\in[T]:E \text{ is true}}\hspace{-4mm} r_{i,t}\leq 2\nu\bar{\beta_T} \sqrt{2MTd \log_\nu\left(1 + \frac{TL^2 }{d\gamma_{\min}}\right)}.
\end{aligned}
\end{equation}
Next, we consider the case where event $E  $ does not hold. The regret in any period between synchronization of length $ t_k=T_k-T_{k-1} $ can be given by
\begin{equation}
\begin{aligned}
&R([T_{k-1},T_{k}]) = \sum_{t=T_{k-1}}^{T_k} \sum_{i=1}^{M} r_{i,t}\\
\leq \;&2\nu\bar{\beta_T} \left( \sum_{i=1}^{M} \sqrt{ t_k \sum_{t=T_{k-1}}^{T_k} \norm{\bm{x}_{i,t}}_{\bm{V}_{i,t}^{-1}}} \right)\leq 2\nu\bar{\beta_T}M  \sqrt{D}
\end{aligned}
\end{equation}
where the last inequality comes from $ t_k \log_\nu\left(  \frac{  \det\left(  \bm{G}_{i,t+t_k} + \gamma_{\max}\textbf{I}\right)}{\det\left( \bm{G}_{i,t}+\gamma_{\min}\textbf{I}  \right)} \right) \leq D   $. Thus, the total regret over all $ T $ rounds where event $ E $ does not hold can be bound as
\begin{equation}\label{ine2}
\begin{aligned}
R(T,\bar{E})& \leq 2R \nu\bar{\beta_T}M  \sqrt{D}\\
& \leq 2\nu\bar{\beta_T}M  \sqrt{D} \left(   d \log_\nu\left( \frac{\gamma_{\max}}{\gamma_{\min}} +\frac{TL^2}{d\gamma_{\min}} \right)+ 1 \right).
\end{aligned}
\end{equation}
Combining inequality \eqref{ine1} and \eqref{ine2}, we have
\begin{equation}
\begin{aligned}
R(T)&=R(T,E) +R(T,\bar{E})\\
& \leq 2\nu\bar{\beta_T} \left(\sqrt{2MTd \log_\nu\left( \frac{\gamma_{\max}}{\gamma_{\min}}+ \frac{TL^2 }{d\gamma_{\min}}\right)} \right.\\
& \left. +  M  \sqrt{D} \left(   d \log_\nu\left( \frac{\gamma_{\max}}{\gamma_{\min}} +\frac{TL^2}{d\gamma_{\min}} \right)+ 1 \right)  \right).
\end{aligned}
\end{equation}
We set $ D=2Td \left(\log_\nu\left( \frac{\gamma_{\max}}{\gamma_{\min}} +\frac{TL^2}{d\gamma_{\min}} \right)+ 1 \right)^{-1} $, we have
\begin{equation}
\begin{aligned}
R(T) &\leq 4\nu \bar{\beta_T}  \sqrt{ 2MTd \log_\nu\left( \frac{\gamma_{\max}}{\gamma_{\min}} + \frac{TL^2 }{d\gamma_{\min}}\right) + 1}.
\end{aligned}
\end{equation}

\end{document}